\documentclass[conference]{IEEEtran}
\IEEEoverridecommandlockouts
\PassOptionsToPackage{linktocpage}{hyperref}
\usepackage{cite}
\usepackage{amsmath,amssymb,amsfonts}
\usepackage{algorithmic}
\usepackage{graphicx}
\usepackage{textcomp}
\usepackage{xcolor}
\usepackage{lipsum} 
\usepackage{hyperref} 
\def\BibTeX{{\rm B\kern-.05em{\sc i\kern-.025em b}\kern-.08em
    T\kern-.1667em\lower.7ex\hbox{E}\kern-.125emX}}
\begin{document}

\title{Open-source High-precision Autonomous Suturing Framework With Visual Guidance\\

\thanks{* is Equal contribution. The first two authors contributed equally.}
\thanks{This work was
supported in part by the CUHK Chow Yuk Ho Technology Centre of Innovative
Medicine, in part by the Multi-Scale Medical Robotics Centre, InnoHk, 8312051, RGC T42-409/18-R, in part by SHIAE (BME-p1-17), and in part by the Natural Science
Foundation of China under Grant U1613202. (Corresponding author: K. W.
Samuel Au)}
\thanks{$^{1}$Hongbin Lin, Bin Li, Yunhui Liu and  Kwok Wai Samuel Au are with Department of Mechanical, The Chinese University of Hong Kong, Hong Kong. {\tt\small \{hongbinlin, binli\}@link.cuhk.edu.hk;  \{yhliu, samuelau\}@cuhk.edu.hk}}
}

\author{Hongbin Lin$^{1,*}$, Bin Li$^{1,*}$, Yunhui Liu$^{1}$ and Kwok Wai Samuel Au$^{1}$
}

\maketitle

\begin{abstract}
Autonomous surgery has attracted increasing attention for revolutionizing robotic patient care, yet remains a distant and challenging goal. In this paper, we propose an image-based framework for high-precision autonomous suturing operation. We first build an algebraic geometric algorithm to achieve accurate needle pose estimation, then design the corresponding keypoint-based calibration network for joint-offset compensation, and further plan and control suture trajectory. Our solution ranked first among all competitors in the AccelNet Surgical Robotics Challenge. Videos and codes can be found in \url{https://sites.google.com/view/accel-2022-cuhk}.

\end{abstract}

\section{Introduction}
Autonomous suturing is a long-standing robotic challenge for surgical autonomy\cite{attanasio2020autonomy}. Yet, research in surgical autonomy was slowed down by limited scopes of benchmarking setups and lack of standardization\cite{munawar2022open}.
Aiming to address these issues, organizers raised the \href{https://collaborative-robotics.github.io/surgical-robotics-challenge/challenge-2021.html}{2021-2022 AccelNet Surgical Robotic Challenge (online)}, providing high-fidelity simulation, standardized problem definitions, and evaluations for benchmarking autonomous suturing\cite{munawar2022open}. Targeting at three main problems in AccelNet Challenge, including needle pose estimation, needle grasp under joint-offset, and the multi-loop suture operation, we built an autonomous suturing framework and achieved state-of-the-art (SOTA) performance.
Our main contribution are:
\begin{itemize}
    \item a SOTA image-guided framework for high-precision autonomous suturing tasks in AccelNet Challenge.
    \item off-the-shelf open-source software of our framework for easy reproduction and future research acceleration in the robotic surgery field. 
\end{itemize}

\section{Method and Result}

\begin{figure*}[!tbp]
  \centering
  \includegraphics[width=1.0\hsize]{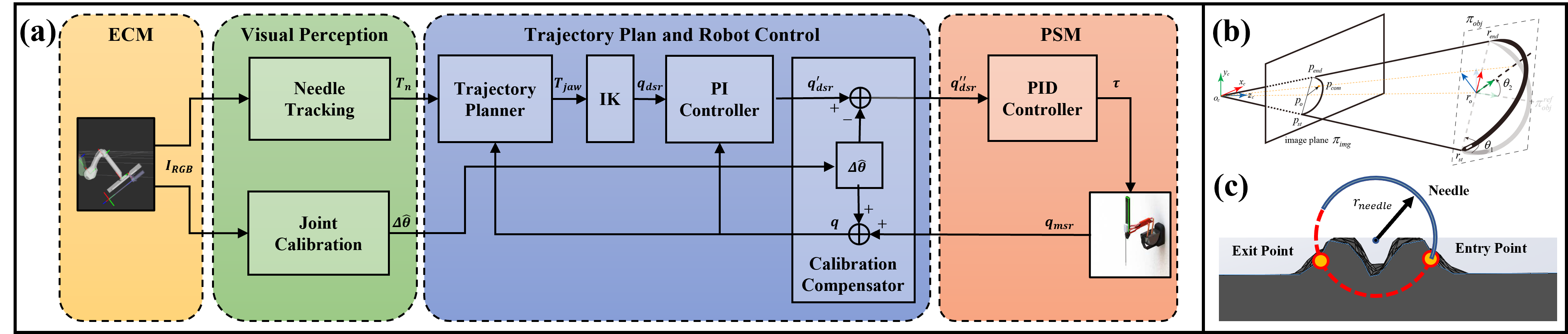}
  \caption{(a) Our proposed framework for autonomous suturing in AccelNet Challenge, (b) disentangled needle physical projection model, and (c) trajectory design for needle insertion and extraction.}
  \vspace{-0.45cm}
  \label{fig:1}
\end{figure*}

\subsection{Visual Perception}
\subsubsection{Needle Tracking}

The goal of this task is to identify the suture pose $\mathbf{T} \in R^{4x4}$ based on stereo visual perception $I_{l,r}$, and it provides the basis for the autonomous suturing operation. To this end, we first build a multi-task network based on MaskRCNN\cite{he2017mask} including the segmentation, and key point heads, which are used to extract coarse needle mask $\mathbf{M}^{l,r}_{c}$ and the start/end key-points $\mathbf{P^{l,r}_{xy}}$, respectively. We further designed a coarse-to-fine strategy to optimize the coarse mask in RGB color space to obtain fine segmentation $\mathbf{M}^{l,r}_{f}$. The above process is formulated as follows:
\begin{equation} 
\label{equ:mask-rcnn}
[\boldsymbol{\mathbf{M}^{l,r}_{c}}, \mathbf{P^{l,r}_{xy}}] = F^{}_{multitask}(I_{l,r}), \boldsymbol{\mathbf{M}^{l,r}_{c}}\xrightarrow[]{optimize}\boldsymbol{\mathbf{M}^{l,r}_{f}}
\end{equation}
To recover the needle pose $\mathbf{T}$ from the extracted mask and key points, we hereby present an algebraic geometry-based algorithm. We define the six Degree of Freedoms (DOFs) of needle $\boldsymbol{x}: [\theta_{1,2}, kp^{st, ed}_{x,y}] \in R^{6}$ to represent the needle pose uniquely which contains four projected coordinates $kp^{st, ed}_{x,y}$ in image plane and two angles $\theta_{1,2}$ locate in the rays-formed plane, shown in Fig. {\ref{fig:1}}b. Given a hypothetical needle pose $\boldsymbol{x}$, the corresponding reprojected axis points set $\mathbf{S}^{(l,r)}(\boldsymbol{x})$ can be calculated. Therefore, the goal of the needle pose estimation is to find a $\boldsymbol{x}$ that minimize the reprojection offset error between points set $\mathbf{S}^{(l,r)}(\boldsymbol{x})$ and needle fine mask $\mathbf{M}^{(l,r)}_{f}$:
\begin{equation}
\label{equ:pose}
 \mathop{\arg\min}\limits_{\boldsymbol{x}} \mathbf{J}_{A} = \sum_{k \in \{r, l\}} \sum_{\boldsymbol{p}_{i} \in \mathbf{M}^{(k)}_f}^{} \mathop{\min}_{\boldsymbol{p}_{j} \in \mathbf{S}^{(k)}(\boldsymbol{x})} || \boldsymbol{p}_i - \boldsymbol{p}_j||_2^2
\end{equation}
Note that the objective $\mathbf{J}_{A}$ can innately handle the partially occluded case, since the occluded part is not reflected in the mask and thus does not affect the optimization process. This feature can further expand the use scenarios for needle pose estimation, since the needle is usually partially occluded by the PSM capture during surgery. The needle pose is optimized in $\boldsymbol{x}$-space with gradient optimization method, and then transformed into Cartesian space $\mathbf{T}$. We randomly placed needle and camera pose with their distance range from 80 to 200 mm, and collected 1k samples for training the multi-task network. The optimization is performed in a maximum of 1.5k steps. Our needle pose estimation algorithm achieved an average position error of 0.3 mm and angular error of 1.1 degrees on the AccelNet suture platform. 
\subsubsection{Joint Calibration with Monocular Camera}

There is an unknown bias $\Delta \boldsymbol{q} \in R^{6}$ between actual and measured joint position, $ \boldsymbol{q} $ and $\boldsymbol{q_{msr}}$, for a Patent Side Manipulator (PSM), where $\Delta \boldsymbol{q}= \boldsymbol{q} - \boldsymbol{q_{msr}}$ . The goal of joint calibration is to identify the unknown joint bias $\Delta \boldsymbol{q}$ during evaluation. First, we tracked $N$ non-colinear feature points on the jaw of PSMs using DeepLabCut (DLC)\cite{mathis2018deeplabcut}, a state-of-the-art, markerless, data-driven framework that achieves high tracking accuracy using limited human-labeled data. $N$ was set to $4$ since $N \geq 3$ suffices to identify a unique pose. The position of pixels for tracked features on a RGB monocular image $\boldsymbol{I_{RBG}}$ along x and y axes, $\boldsymbol{x_{ft}}$ and $\boldsymbol{y_{ft}}$ respectively, were predicted by a trained deep-neural-network function $f_{dlc}$ using DLC, which can be formulated as:
\begin{equation} 
\label{equ:dlc}
 \begin{bmatrix}
 \boldsymbol{x_{ft}}\\
\boldsymbol{y_{ft}}
 \end{bmatrix}=f_{dlc}(\boldsymbol{I_{RBG}}), 
\end{equation}
We define a pose estimation function $f_{pose}$, where a unique solution of the pose of the jaw $T_{jaw}$ can be calculated by $T_{jaw}=f_{pose}(\boldsymbol{x_{ft}},\boldsymbol{y_{ft}})$. Since the DOFs of a PSM is $6$, there will be either a single solution or multiple solutions of inverse kinematics $f_{ik}$ in a discrete solution set $\{\boldsymbol{x}: \boldsymbol{x}=f_{ik}(T_{jaw})\}$ for each jaw pose $T_{jaw}$ within the reachable workspace. Since $\boldsymbol{q} = \boldsymbol{q_{msr}} + \Delta \boldsymbol{q}$, we can reduce the number of possible solutions by constraining the solution set as $\{\boldsymbol{x}: \boldsymbol{x}=f_{ik}(T_{jaw}), \|\boldsymbol{x}-\boldsymbol{q_{msr}} \| \leq \|\Delta \boldsymbol{q}\| \}\}$. Via numerical sampling, we found some $\boldsymbol{q_{msr}}$ so that there is only a single solution in the constrained solution set for PSM under the test condition of AccelNet Challenge ($\|\Delta \boldsymbol{q}\| \leq 10^{\circ}$).
For those $\boldsymbol{q_{msr}}$, a unique $\Delta \boldsymbol{q}$ is formulated as:
\begin{equation} 
\label{equ:dlc}
\Delta \boldsymbol{q}=\boldsymbol{q}-\boldsymbol{q_{msr}}=
f_{ik}(f_{pose}( \begin{bmatrix}
 \boldsymbol{x_{ft}}\\
\boldsymbol{y_{ft}}
 \end{bmatrix}))-\boldsymbol{q_{msr}}=f_{cal}(\begin{bmatrix}
 \boldsymbol{x_{ft}}\\
\boldsymbol{y_{ft}}
 \end{bmatrix}), 
\end{equation}
where $f_{cal}$ is defined as calibration function. Since both $f_{ik}$ and $f_{pose}$ are error-prone and computational expensive, we used Multi-Layer Perceptron (MLP) to approximate the calibration function $f_{cal}$. PSMs were moved to $10k$ configurations by a positional controller, where $\Delta \boldsymbol{q}$ was randomly sampled within the testing range $[-5^{\circ}, 5^{\circ}]$ and the desired configuration was set to $\boldsymbol{q_{msr}}$ that satisfies aforementioned single-solution claim. MLP has 3 hidden layers with 400,300 and 200 neurons respectively followed by ReLU \cite{nair2010rectified} activation function except the last layer. Input and output signals of the MLP were preprocessed using scalers of normal distribution similar to \cite{lin2021learning}. Adam \cite{kingma2014adam} was used to optimize the model. Our calibration approach achieved $0.2\pm0.2$ mean error for each joint of PSMs.

\subsection{Robot Control}
We designed a control strategy (including trajectory planning, IK (Inverse Kinematics), PI Controller, and calibration compensator) to servo PSMs for suturing tasks.  
\subsubsection{Trajectory Planning}
For needle insertion and extraction, waypoints were sampled from a needle-radius circular trajectory passing through the entry and exit suturing ports (See Fig. \ref{fig:1}c). Alternatively, waypoints were automatically generated using linear interpolation given the current and desired poses of PSMs for other cases.  

\subsubsection{IK}
We used the analytical IK engine provided by the AccelNet challenge since it is less computationally expensive and more stable compared to other numerical methods.

\subsubsection{PI Controller}
Significant error (about $1.5^{\circ}$ for revolute joints and $0.1 mm$ for prismatic joints) between desired and measured joint positions without considering the unknown joint bias was observed using a low-level PID controller in the test condition of AccelNet Challenge. Therefore, We designed a high-level PI controller to minimize the error. The error was reduced by $98\%$ after adding our high-level controller.

\subsubsection{Calibration Compensator}
A joint offset $\Delta \boldsymbol{\hat{q}}$ estimated by our joint calibration method was further used for compensation. Both measured and desired join positions were compensated as Fig. \ref{fig:1}a shows.

\section{Conclusion}
In this paper, we proposed an autonomous suturing framework for AccelNet surgical robotics challenge. Our framework includes algebraic geometric needle pose estimation, keypoint-based joint calibration, robotic planning and control for autonomous suturing. Future research direction includes real-time needle and tool tracking, deformable tissue manipulation during needle insertion and extraction, and collision avoidance in a complex confined dynamic environment.


\bibliographystyle{IEEEtran}
\bibliography{main}

\begin{thebibliography}{1}
\providecommand{\url}[1]{#1}
\csname url@samestyle\endcsname
\providecommand{\newblock}{\relax}
\providecommand{\bibinfo}[2]{#2}
\providecommand{\BIBentrySTDinterwordspacing}{\spaceskip=0pt\relax}
\providecommand{\BIBentryALTinterwordstretchfactor}{4}
\providecommand{\BIBentryALTinterwordspacing}{\spaceskip=\fontdimen2\font plus
\BIBentryALTinterwordstretchfactor\fontdimen3\font minus
  \fontdimen4\font\relax}
\providecommand{\BIBforeignlanguage}[2]{{%
\expandafter\ifx\csname l@#1\endcsname\relax
\typeout{** WARNING: IEEEtran.bst: No hyphenation pattern has been}%
\typeout{** loaded for the language `#1'. Using the pattern for}%
\typeout{** the default language instead.}%
\else
\language=\csname l@#1\endcsname
\fi
#2}}
\providecommand{\BIBdecl}{\relax}
\BIBdecl

\bibitem{attanasio2020autonomy}
A.~Attanasio \emph{et~al.}, ``Autonomy in surgical robotics,'' \emph{Annual
  Review of Control, Robotics, and Autonomous Systems}, vol.~4, pp. 651--679,
  2021.

\bibitem{munawar2022open}
A.~Munawar \emph{et~al.}, ``Open simulation environment for learning and
  practice of robot-assisted surgical suturing,'' \emph{IEEE Robotics and
  Automation Letters}, vol.~7, no.~2, pp. 3843--3850, 2022.

\bibitem{he2017mask}
K.~He, G.~Gkioxari, P.~Doll{\'a}r, and R.~Girshick, ``Mask r-cnn,'' pp.
  2961--2969, 2017.

\bibitem{mathis2018deeplabcut}
A.~Mathis, P.~Mamidanna, K.~M. Cury, T.~Abe, V.~N. Murthy, M.~W. Mathis, and
  M.~Bethge, ``Deeplabcut: markerless pose estimation of user-defined body
  parts with deep learning,'' \emph{Nature neuroscience}, vol.~21, no.~9, pp.
  1281--1289, 2018.

\bibitem{nair2010rectified}
V.~Nair and G.~E. Hinton, ``Rectified linear units improve restricted boltzmann
  machines,'' pp. 807--814, 2010.

\bibitem{lin2021learning}
H.~Lin, Q.~Gao, X.~Chu, Q.~Dou, A.~Deguet, P.~Kazanzides, and K.~S. Au,
  ``Learning deep nets for gravitational dynamics with unknown disturbance
  through physical knowledge distillation: Initial feasibility study,''
  \emph{IEEE Robotics and Automation Letters}, vol.~6, no.~2, pp. 2658--2665,
  2021.

\bibitem{kingma2014adam}
D.~P. Kingma \emph{et~al.}, ``Adam: A method for stochastic optimization,''
  \emph{International Conference on Learning Representations ({ICLR})}, 2015.

\end{thebibliography}

\end{document}